\documentclass[11pt,a4paper]{article}
\usepackage[margin=1in]{geometry}
\usepackage[utf8]{inputenc}
\usepackage[T1]{fontenc}
\usepackage{lmodern}
\usepackage{microtype}
\usepackage{amsmath,amssymb,amsthm}
\usepackage{booktabs}
\usepackage{tikz}
\usetikzlibrary{arrows.meta,positioning,calc,shapes}
\usepackage{pgfplots}
\pgfplotsset{compat=1.18}
\usepackage[backend=biber,style=numeric-comp,sorting=nyt]{biblatex}
\renewbibmacro*{in:}{}
\DeclareFieldFormat{parens}{#1}  

\usepackage[hidelinks]{hyperref}
\usepackage{cleveref}
\usepackage{xcolor}
\usepackage{array}
\usepackage{caption}
\usepackage{url}
\usepackage[strings]{underscore} 

\definecolor{ActuBlue}{HTML}{0E4A8A}
\definecolor{ActuGold}{HTML}{C9A227}
\definecolor{ActuTeal}{HTML}{2C7A7B}
\definecolor{ActuRed}{HTML}{B53F3F}
\definecolor{ActuGreen}{HTML}{2E7D32}
\definecolor{ActuLight}{HTML}{E5F0FB}

\providecommand{\mathdefault}[1][]{}

\title{ActuBench: A Multi-Agent LLM Pipeline for Generation and\\Evaluation of Actuarial Reasoning Tasks}
\author{Jan-Philipp Schmidt\\TH K\"oln, Institut f\"ur Versicherungswesen (ivwK\"oln)\\\texttt{jan-philipp.schmidt@th-koeln.de}}
\date{\today}

\begin{document}
\maketitle

\begin{abstract}
\noindent
We present ActuBench, a multi-agent LLM pipeline for the automated generation and evaluation of advanced actuarial assessment items aligned with the International Actuarial Association (IAA) Education Syllabus. The pipeline separates four LLM roles by adapter: one agent drafts items, one constructs distractors, a third independently verifies both stages and drives bounded one-shot repair loops, and a cost-optimized auxiliary agent handles Wikipedia-note summarization and topic labelling. The items, per-model responses and complete leaderboard are published as a browsable web interface at \url{https://actubench.de/en/}, allowing readers and practitioners to inspect individual items without a repository checkout. We evaluate 50 language models from eight providers on two complementary benchmarks --- 100 empirically hardest multiple-choice items and 100 open-ended items scored by an LLM judge --- and report three headline findings. First, multi-agent verification is load-bearing: the independent verifier flags a majority of drafted items on first pass, most of which the one-shot repair loop resolves. Second, locally-hosted open-weights inference sits on the cost--performance Pareto front: a Gemma~4 model running on consumer hardware and a Cerebras-hosted 120B open-weights model dominate the near-zero-cost region, with the latter within one item of the top of the leaderboard. Third, MCQ and LLM-as-Judge rankings differ meaningfully: the MCQ scaffold inflates the performance ceiling, and Judge-mode evaluation is needed to discriminate at the frontier.
\end{abstract}

\section{Introduction}
\label{sec:intro}

Actuarial science is one of the most technically demanding professional domains: it requires precise mathematical reasoning under stated assumptions, deep familiarity with insurance and financial theory, sensitivity to regulatory frameworks, and the ability to trade off computation, approximation and model risk. Internationally, the body of knowledge expected of a qualified actuary is codified by the International Actuarial Association (IAA) Education Syllabus,\footnote{\url{https://actuaries.org/app/uploads/2025/06/20250525_EducationSyllabus_Final.pdf}} which covers life insurance, non-life insurance, pensions, health insurance, enterprise risk management, and actuarial data science through more than a hundred learning objectives. These objectives underwrite the qualification examinations of national actuarial bodies such as the Deutsche Aktuarvereinigung (DAV), the Society of Actuaries (SOA) and the Institute and Faculty of Actuaries (IfoA).

Constructing advanced assessment material for this body of knowledge --- whether multiple-choice items, computational problems, or open-ended reasoning tasks --- is expensive. Skilled subject-matter experts must craft questions that are unambiguous, internally consistent, grounded in current knowledge, and calibrated for difficulty; the process is slow even for a single examination committee. Meanwhile, language models have made rapid progress on general and domain-specific reasoning benchmarks~\cite{hendrycks2021math,phan2025hle,naveed2024overview}, raising the question of how capable the current frontier is on actuarial tasks specifically and what it would take to measure that capability reproducibly. Existing actuarial-LLM studies either survey use-cases without benchmarking~\cite{balona2023actuarygpt}, target a narrow quantitative subdomain~\cite{hao2025reinsurance}, or benchmark Chinese-language insurance knowledge~\cite{zhou2025cufeinse}; to our knowledge no English-language LLM benchmark aligned with the IAA Education Syllabus exists.

We address this with ActuBench: a multi-agent LLM pipeline that generates IAA-aligned assessment items from learning objectives and Wikipedia-grounded notes, verifies them with an independent verifier agent, and evaluates language models on the resulting pool in two complementary modes. Our contributions are fivefold.

\begin{enumerate}
  \item \textbf{A multi-agent generation pipeline with an independent verifier.} Four LLM roles are distinguished by adapter: Agent~A drafts items, Agent~B constructs distractors, Agent~C independently verifies both, and a cost-optimized auxiliary agent handles Wikipedia-note summarization and topic-label assignment. Agent~C never contributes generated content; it drives bounded repair loops triggered by its verdicts. This verifier-first role separation is the pipeline's principal methodological differentiator (\cref{sec:pipeline:agentc}).
  \item \textbf{A dual evaluation protocol: MCQ and LLM-as-Judge.} We evaluate each model both in a classical four-option MCQ setting and in an open-ended setting where a judge LLM scores the evaluatee's free-text response against the known correct answer. The two views produce meaningfully different rankings of the same models (\cref{sec:results:mvj}).
  \item \textbf{A live, web-accessible benchmark viewer} at \url{https://actubench.de/en/}. The 100 MCQ items, 100 Judge items, and every model response are browsable and searchable through a web UI, allowing readers and practitioners to inspect individual items and per-model answers without a GitHub checkout.
  \item \textbf{An empirical first evaluation of 50 language models.} We evaluate 50 models spanning eight providers --- Anthropic, Google, OpenAI, xAI, DeepSeek, Mistral, Cohere, and open-weights hosted endpoints including locally-run Gemma --- on 100 empirically hardest MCQ items and 100 open-ended Judge items (\cref{sec:methodology:sampling}).
  \item \textbf{Evidence that locally-hosted open-weights inference is on the cost--performance Pareto front.} A Cerebras-hosted open-weights 120B-parameter model reaches 97\,\% MCQ accuracy on the hardest items at near-zero cost; a Gemma~4 model running on a consumer GPU via Ollama reaches 85\,\% at zero marginal cost. At the paid end of the scale, per-answer frontier accuracy can be obtained at an order-of-magnitude lower cost than from the flagship reasoning-mode variants (\cref{sec:results:pareto}).
\end{enumerate}

The remainder of the paper is structured as follows. \Cref{sec:related} reviews related work. \Cref{sec:pipeline} describes the generation pipeline. \Cref{sec:methodology} specifies the two benchmark modes and the item-selection protocol. \Cref{sec:results} reports empirical results along three axes: cost--performance, sector-level accuracy, and MCQ-vs-Judge divergence. \Cref{sec:discussion} discusses reasoning-mode variants, Agent-C repair statistics, and implications for practical LLM use in actuarial settings. \Cref{sec:limitations} and~\cref{sec:conclusion} close.

\section{Background and Related Work}
\label{sec:related}

ActuBench sits at the intersection of six research threads: automatic question generation, LLM benchmark design, LLM-as-judge evaluation, multi-agent LLM pipelines, benchmark contamination, and the still-thin literature on LLMs in actuarial science.

\subsection{Automatic Question Generation}
\label{sec:related:aqg}

Automatic question generation (AQG) has been studied for over two decades. Early work on assessment-oriented AQG used syntactic transformations on parse trees combined with statistical ranking, exemplified by \textcite{heilman2010good}'s seminal factual-QG system. Neural sequence-to-sequence models extended this line to learnt question generators, with \textcite{du2017learningtoask} training a seq2seq model on sentence-question pairs from SQuAD (optionally conditioned on paragraph context). \textcite{kurdi2020survey} provide the authoritative systematic review of AQG for educational purposes across rule-based, template, statistical and neural eras. With instruction-tuned LLMs, generating domain multiple-choice questions with correct answers, plausible distractors and rationales became feasible at scale; \textcite{elkins2023useful} evaluated InstructGPT-generated educational questions in a teacher-expert study and found high relevance, grammaticality and answerability, with most items requiring only minor or no edits. That result sharpens the remaining quality-control question: not whether LLMs can draft plausible items, but whether automated verification can catch the residual factual and difficulty-calibration errors without teacher review. ActuBench's Agent-A/Agent-B split with an independent Agent-C verifier is a direct answer to that question.

\subsection{LLM Benchmarks}
\label{sec:related:benchmarks}

A broad family of LLM benchmarks covers general academic knowledge (MMLU~\cite{naveed2024overview}), mathematical reasoning (MATH~\cite{hendrycks2021math}, DROP~\cite{dua2019drop}), and cross-disciplinary graduate-level reasoning (HLE~\cite{phan2025hle}). Domain-specific benchmarks include FinQA for numerical reasoning on financial documents~\cite{chen2021finqa} and various medical-licensing datasets. A recurring concern across this literature is that static benchmarks saturate quickly as models improve, motivating designs that retain only the items current models most frequently miss~\cite{phan2025hle}; our empirically-hardest-100 construction (\cref{sec:methodology:sampling}) adopts this idea.

\subsection{LLM-as-Judge Evaluation}
\label{sec:related:judge}

The LLM-as-judge paradigm was popularized by \textcite{zheng2023mtbench} (MT-Bench and Chatbot Arena), who demonstrated that capable LLMs, particularly GPT-4, can approximate human preference scoring on open-ended tasks at above 80\,\% agreement. \textcite{kim2023prometheus} trained Prometheus, an open-source 13-billion-parameter fine-tuned evaluator LLM, to produce fine-grained rubric-grounded verdicts with reference answers. Critical follow-up work documented systematic biases: \textcite{wang2023unfair} show that LLM judges exhibit pronounced positional bias in the sense that simply reordering candidate responses in the prompt can flip the quality ranking. For domains where a ground-truth correct answer exists (as in actuarial exam items), the judge's task is narrower than in preference-style evaluation: the judge must decide whether an extracted answer matches a known correct answer rather than rank competing responses. ActuBench's judge prompt (\cref{sec:methodology:judge}) enforces this framing explicitly.

\subsection{Multi-Agent LLM Pipelines}
\label{sec:related:multiagent}

Multi-agent LLM systems partition a task across roles with distinct prompts and, increasingly, distinct underlying models. \textcite{hong2023metagpt} (MetaGPT) formalizes role specialization for software-engineering tasks; \textcite{wu2023autogen} (AutoGen) provides a domain-agnostic framework for multi-agent LLM conversation, with demonstrations across mathematics, retrieval-augmented generation, optimization, chess and coding. \textcite{chen2023agentverse} demonstrate emergent collaboration in open-ended problem solving. \textcite{du2023multiagent_debate} introduce multi-agent debate as a mechanism to reduce hallucinations and improve factuality; the empirical pattern they report --- agents that individually err but converge to correct answers under mutual critique --- implicitly supports the more general verification-generation asymmetry we rely on in \cref{sec:pipeline:agentc}. ActuBench applies the same role-specialization principle but restricts the scope to a single, closed content-generation task: one adapter drafts, another audits distractors, a third independently verifies. The restriction yields a well-scoped, bounded-cost pipeline in which each agent's prompt is short and each verdict auditable --- distinct from the open-ended negotiation of debate-style systems.

\subsection{Benchmark Contamination}
\label{sec:related:contamination}

A fast-growing literature documents data contamination as a central threat to static LLM benchmarks: popular benchmark items often appear in pre-training corpora and model performance on them may not reflect genuine generalization~\cite{sainz2023contamination,deng2024tsguessing,zhou2023cheater,xu2024contamination_survey}. Our design addresses this in two ways: the pipeline can generate fresh items on demand from new learning objectives, and the item pool released at \url{https://actubench.de/en/} is a curated subset rather than a static dump.

\subsection{Actuarial AI}
\label{sec:related:actuarialai}

Actuarial AI has historically focused on classical machine learning on tabular and time-series data: Richman's surveys~\cite{richman2018ai,richman2024vision} and \textcite{wuthrich2023statistical}'s textbook treat deep learning for mortality, reserving and non-life pricing, while \textcite{troxler2022nlp} demonstrate transformer-based NLP for claims descriptions. LLM-specific studies have only recently appeared. \textcite{balona2023actuarygpt} is framed as a practitioner survey of LLM use-cases in insurance rather than an evaluation benchmark. \textcite{hao2025reinsurance} build quantitative QA pairs from reinsurance training materials and fine-tune Llama-2 variants; their scope is narrow and their item set non-public. The concurrent Chinese-language CUFEInse benchmark~\cite{zhou2025cufeinse} evaluates eleven LLMs on 14{,}430 insurance-operations questions spanning theory, industry, safety, agents and logical rigor. CUFEInse and ActuBench are complementary rather than overlapping: CUFEInse targets insurance operations on Chinese-language industry content at large scale, while ActuBench targets IAA-syllabus-aligned actuarial reasoning on English-language content with a smaller but methodologically controlled item set and an independent-verifier generation pipeline. To our knowledge, ActuBench is the first English-language actuarial LLM benchmark aligned with the IAA Education Syllabus.

\section{Generation Pipeline}
\label{sec:pipeline}

\subsection{Overview}
\label{sec:pipeline:overview}

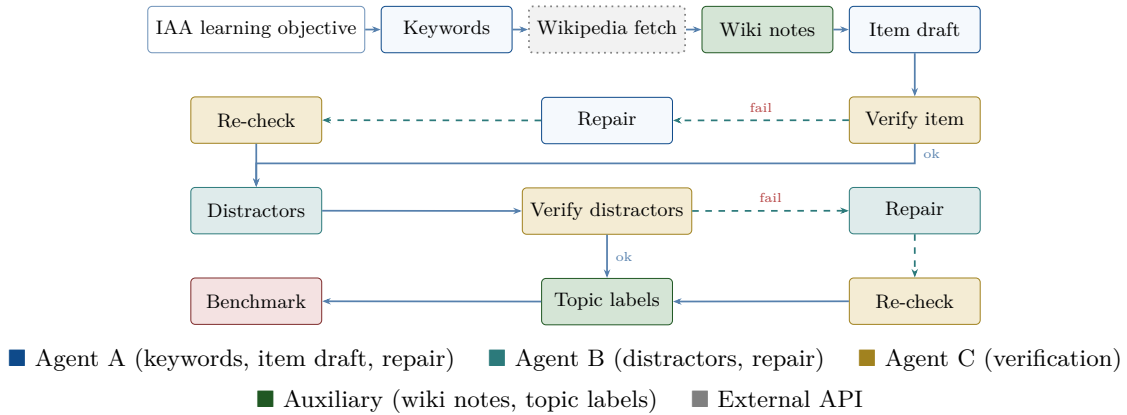
\begin{figure}[!htbp]
\centering
\scalebox{0.82}{%
\begin{tikzpicture}[
  node distance=0.45cm and 0.25cm,
  every node/.style={font=\footnotesize},
  agentA/.style={draw=ActuBlue, rounded corners=2pt, fill=ActuLight!40, align=center, minimum height=0.75cm, minimum width=2.1cm},
  agentB/.style={draw=ActuTeal, rounded corners=2pt, fill=ActuTeal!15, align=center, minimum height=0.75cm, minimum width=2.1cm},
  agentC/.style={draw=ActuGold!80!black, rounded corners=2pt, fill=ActuGold!20, align=center, minimum height=0.75cm, minimum width=2.1cm},
  aux/.style={draw=ActuGreen!70!black, rounded corners=2pt, fill=ActuGreen!20, align=center, minimum height=0.75cm, minimum width=2.1cm},
  apibox/.style={draw=gray, rounded corners=2pt, fill=gray!10, align=center, minimum height=0.75cm, minimum width=2.1cm, thick, dotted},
  startbox/.style={draw=ActuBlue!60, rounded corners=2pt, fill=white, align=center, minimum height=0.75cm, minimum width=2.1cm},
  arr/.style={-{Stealth[length=4pt]}, thick, ActuBlue!70},
  rarr/.style={-{Stealth[length=4pt]}, thick, ActuTeal, dashed},
]

  \node[startbox] (lo) {IAA learning objective};
  \node[agentA, right=of lo] (kw) {Keywords};
  \node[apibox, right=of kw] (wiki) {Wikipedia fetch};
  \node[aux, right=of wiki] (notes) {Wiki notes};
  \node[agentA, right=of notes] (draft) {Item draft};

  \node[agentC, below=0.7cm of draft] (c1) {Verify item};
  \node[agentA, below=0.7cm of wiki] (a4) {Repair};
  \node[agentC, below=0.7cm of lo] (c1r) {Re-check};

  \node[agentB, below=0.7cm of c1r] (b1) {Distractors};
  \node[agentC, below=0.7cm of a4] (c2) {Verify distractors};
  \node[agentB, below=0.7cm of c1] (b1r) {Repair};
  \node[aux, below=0.7cm of c2] (b2) {Topic labels};
  \node[agentC, below=0.7cm of b1r] (c2r) {Re-check};

  \node[draw=ActuRed!70!black, rounded corners=2pt, fill=ActuRed!15, align=center, minimum height=0.75cm, minimum width=2.1cm, below=0.7cm of b1] (bench) {Benchmark};

  \draw[arr] (lo) -- (kw);
  \draw[arr] (kw) -- (wiki);
  \draw[arr] (wiki) -- (notes);
  \draw[arr] (notes) -- (draft);
  \draw[arr] (draft) -- (c1);

  \draw[arr] (c1) -- node[right, font=\tiny] {ok} ++(0,-0.7cm) -| (b1);
  \draw[rarr] (c1) -- node[above, font=\tiny] {\textcolor{ActuRed}{fail}} (a4);
  \draw[rarr] (a4) -- (c1r);
  \draw[arr] (c1r) -- (b1);

  \draw[arr] (b1) -- (c2);
  \draw[arr] (c2.south) -- node[right, font=\tiny] {ok} (b2.north);
  \draw[rarr] (c2) -- node[above, font=\tiny] {\textcolor{ActuRed}{fail}} (b1r);
  \draw[rarr] (b1r.south) -- (c2r);
  \draw[arr] (c2r) -- (b2);

  \draw[arr] (b2) -- (bench);

\end{tikzpicture}%
}

\vspace{0.4em}
{\footnotesize
\textcolor{ActuBlue}{$\blacksquare$} Agent~A (keywords, item draft, repair) \quad
\textcolor{ActuTeal}{$\blacksquare$} Agent~B (distractors, repair) \quad
\textcolor{ActuGold!80!black}{$\blacksquare$} Agent~C (verification)\\[0.2em]
\textcolor{ActuGreen!70!black}{$\blacksquare$} Auxiliary (wiki notes, topic labels) \quad
\textcolor{gray}{$\blacksquare$} External API
}

\caption{ActuBench generation pipeline. Four LLM roles plus an external Wikipedia API cooperate to produce one assessment item. Agent~A drafts (blue); Agent~B builds distractors (teal); Agent~C verifies both independently (gold); a cost-optimized auxiliary handles note summarization and topic labelling (green). Solid arrows are the primary flow; dashed red-labeled arrows are repair branches taken only on verification failure.}
\label{fig:workflow}
\end{figure}

ActuBench processes each assessment item through a linear sequence of stages, each of which persists its output to the database before the next stage begins. Figure~\ref{fig:workflow} shows the full workflow. Four distinct LLM roles are distinguished, each bound to its own adapter and potentially its own underlying model:

\begin{itemize}
  \item \textbf{Agent~A} drafts content (keyword extraction, item draft, and --- when required --- item repair). Agent~A is typically the strongest and most expensive adapter in the pipeline, and in the generation run used for this paper it was bound to a reasoning-mode model (\texttt{claude-sonnet-4-6:thinking}).
  \item \textbf{Agent~B} specializes in distractor design and distractor repair. Agent~B does not need the deeper reasoning of Agent~A --- the task is more constrained and creative rather than multi-step --- so in the generation run used here it was bound deliberately to a non-reasoning adapter (\texttt{claude-sonnet-4-6}).
  \item \textbf{Agent~C} is the independent verifier. It performs two dedicated verification stages (once on the item stem and correct answer, once on the four-option ensemble) and never contributes generated content. Agent~C is bound to its own adapter, allowing a separate, potentially more critical model to be used for verification (here \texttt{o3:reasoning}).
  \item The \textbf{auxiliary agent} handles all cost-sensitive ancillary work: summarizing Wikipedia extracts into structured notes, and later assigning topic labels over the actuarial sectors. This adapter is always a cheap, fast model (here \texttt{gpt-5.4-mini}); the cost savings of keeping these two stages off the primary generator add up across hundreds of items.
\end{itemize}

Generation proceeds top-to-bottom in Figure~\ref{fig:workflow}. A learning objective from the International Actuarial Association (IAA) Education Syllabus is turned into a Wikipedia search query by Agent~A; relevant article excerpts are fetched through the MediaWiki API; the auxiliary agent summarizes them into bullet-point notes; Agent~A uses the notes plus the learning objective to draft an item. Agent~C then verifies the item and, on a failure verdict, Agent~A produces a single-shot repair that is re-checked. Agent~B then constructs three distractors; Agent~C verifies those and, on failure, triggers a single-shot Agent~B repair with its own re-check. Finally the auxiliary agent assigns topic labels to the finished item and the item enters the evaluation pool.

\subsection{Agent~A}
\label{sec:pipeline:agenta}

Agent~A carries the two stages that benefit most from strong underlying reasoning: item drafting and repair. Each stage emits a structured JSON output that is parsed and cached.

\textbf{Keyword extraction.} Given an IAA learning objective (code, textual description, subject and topic), Agent~A produces a small JSON object with primary search terms, synonyms and abbreviations, disambiguation exclusions, and between one and three candidate search strings. Requiring explicit disambiguation-guard terms (e.g., ``exclude: option pricing in finance'' when the objective is on actuarial option costs) sharpens downstream Wikipedia retrieval measurably.

\textbf{Item drafting.} From the learning-objective text, the sampled difficulty archetype, and the Wikipedia notes produced by the auxiliary agent, Agent~A produces the question stem, the single correct answer, and a detailed rationale justifying why that answer is correct. The prompt enforces several hard constraints: English-language only, single best answer, no multiple defensible answers under reasonable interpretation, explicit statement of any assumption needed for uniqueness, and no reliance on parametric knowledge that cannot be traced to the Wikipedia notes. The correct answer occupies a fixed internal slot by construction; randomization into an MCQ-presentation order is done only at evaluation time.

\textbf{Repair.} When Agent~C returns a non-ok verdict on the item, Agent~A is re-invoked with the item draft and Agent~C's failure report, and produces a revised item that attempts to resolve the flagged issue. This is a single-shot repair, not a negotiation: the revised item then goes through a re-check, and if that also fails the item is persisted with a failure flag but not discarded.

\subsection{Agent~B}
\label{sec:pipeline:agentb}

Agent~B specializes in distractor design. Given the question stem, the correct answer with its rationale, and the Wikipedia notes, Agent~B produces three distractors. Each distractor is required to be incorrect given the same notes and assumption set that make~A correct, individually plausible to a learner at the target difficulty, and distinguishable from the other distractors. Each distractor is accompanied by a short free-text rationale describing the specific misconception or computational slip it is meant to probe. This prompt design follows the distractor-as-misconception-probe tradition in educational assessment~\cite{heilman2010good,kurdi2020survey}, adapted to the LLM-generation setting. On Agent~C's failure verdict for the distractor ensemble, Agent~B is re-invoked once with the failure report and produces a revised set, which is then re-checked.

\subsection{Agent~C --- Independent Verification}
\label{sec:pipeline:agentc}

Agent~C is the principal methodological differentiator of ActuBench. It performs two verification stages --- one on the item stem and correct answer, one on the full four-option ensemble --- and returns an ok/fail verdict with a structured failure report. Unlike Agents~A and~B, Agent~C never contributes generated content; it only consumes a completed draft. A separate LLM adapter is bound to Agent~C so that a different model can be used for verification than for generation. This role and adapter separation was driven by two converging observations.

First, there is evidence that LLMs can be more accurate at judging a candidate solution than at producing one: \textcite{zheng2023mtbench} show that strong LLMs agree with human judges on open-ended tasks at above 80\,\%, and the convergence pattern documented in multi-agent debate~\cite{du2023multiagent_debate} --- where individually-erring agents converge to correct answers under mutual critique --- is consistent with the same asymmetry. This is the same asymmetry exploited in retrieval-augmented generation, multi-agent debate, and outcome-supervised training: it is usually easier to recognize a flaw than to avoid one. For item construction, the implication is direct: having the \emph{same} agent both draft an item and declare it correct is prone to shared blind spots, while routing verification through a model that has not seen the draft-in-progress breaks this shared-failure mode. Empirically, Agent~C flags more than three in five items Agent~A produced in our run as requiring item-level repair (\cref{sec:discussion:agentc}), which is substantially higher than any self-verification rate we observed in preliminary experiments.

Second, verification is a well-scoped task with structured output, for which smaller or more reasoning-oriented models often outperform larger general-purpose generators in our domain. Decoupling the verifier adapter from the generator lets us pair a strong generator with an efficient verifier, and lets us swap either component without restructuring the pipeline. The role separation is analogous to the generator--critic split in multi-agent coding frameworks~\cite{hong2023metagpt,wu2023autogen}, specialized here to content verification on a single generation task rather than to open-ended problem solving.

\textbf{Item verification.} Given the learning objective, the difficulty archetype, the Wikipedia notes, the question stem, the correct answer, and the correct-answer rationale, Agent~C returns a verdict \texttt{ok}$\,\in\,\{$true, false$\}$ with a structured failure report on \texttt{ok=false}. The report enumerates: ambiguity (can another interpretation of the stem make another answer defensible?), factual error (does the correct answer conflict with the notes?), and underspecification (is a required assumption missing?). This report is both stored in the database and passed to Agent~A's repair stage as the prompt context.

\textbf{Distractor verification.} Given all four options with their rationales, Agent~C checks that the distractors are non-duplicative, individually plausible, individually incorrect under the item's stated assumption set, and probe distinct misconceptions. On failure, the report identifies which distractor(s) are problematic and for what reason, and is consumed by Agent~B's repair stage to produce a revised set.

\textbf{Repair loop structure.} Each verification stage admits at most one repair iteration. The re-check that follows each repair is final: if it still fails, the item is persisted with the failure flag and continues to the topic-label stage. This bounded-retry design trades a modest yield loss (about 15\,\% of items in our run were still flagged after repair; see \cref{sec:discussion:agentc}) for a predictable upper bound on per-item cost.

\subsection{Auxiliary Agent}
\label{sec:pipeline:aux}

The auxiliary agent handles two cost-sensitive ancillary stages: Wikipedia-note summarization (after the MediaWiki fetch) and topic-label classification (after the item is finalized). The auxiliary adapter is always bound to a cheap, fast model: the two stages are well-structured and do not benefit from frontier reasoning, and the savings compound over hundreds of generated items. Note summarization produces a fixed bullet-list schema grouping extract content into named facts with provenance; topic labelling assigns each item zero or more of six actuarial-sector labels (life, non-life, AFIR-ERM, ADS-AI, health, pension). The labels are not mutually exclusive --- an item on longevity-linked non-life hybrid products may carry both \emph{Life} and \emph{Non-life} --- and feed the sector-level analysis in \cref{sec:results:sectors}.

\section{Benchmark Methodology}
\label{sec:methodology}

We evaluate language models on the items produced by the pipeline of \cref{sec:pipeline} in two complementary modes. The two benchmarks operate on \emph{disjoint} item sets drawn from the same generation pipeline: 100 items for the MCQ benchmark and 100 different items for the Judge benchmark, so that no model sees the same item in both evaluation modes.

\subsection{MCQ Evaluation}
\label{sec:methodology:mcq}

In MCQ mode, the evaluated model is shown the question stem together with the four options A--D and is instructed to respond with a single letter. A deterministic per-run pseudo-random shuffle decides, for each item, which of the four options receives label A, B, C, or D at presentation time (the correct answer is stored internally in a fixed slot and never appears in the same letter position across runs). The shuffle map is persisted with every answer row for later analysis. The prompt is deliberately terse to minimize format-following confounds:
\begin{quote}\small
\emph{System:} You are taking a multiple-choice actuarial exam. Select the single best answer. Respond with ONLY the letter A, B, C, or D --- no explanation, no punctuation.

\emph{User:} Question: \texttt{\{stem\}} \quad A) \dots B) \dots C) \dots D) \dots
\end{quote}
The response is parsed by extracting the first token in $\{\texttt{A},\texttt{B},\texttt{C},\texttt{D}\}$; if no such token is present the answer is recorded as unparseable and counted as incorrect. Token counts and per-call cost are computed from the provider's reported usage and the pricing table at call time.

\subsection{Judge Evaluation}
\label{sec:methodology:judge}

Judge mode removes the multiple-choice scaffold entirely. The evaluated model sees only the question stem and is asked to answer in free text, showing its reasoning and then stating its final answer:
\begin{quote}\small
\emph{Evaluatee system:} You are taking an actuarial exam. Answer the following question as accurately and concisely as possible. Show your reasoning briefly, then state your final answer clearly.
\end{quote}
A second LLM --- the \emph{judge} --- then assesses the free-text response against the known correct answer. The judge prompt returns a structured JSON verdict with four fields: the final answer extracted from the response, a short reasoning note, a yes/no correctness verdict, and a 0--100 confidence score (the evaluatee's self-reported confidence where available, 100 otherwise). The judge's instruction restricts its scope to comparing the extracted answer to the correct answer on \emph{meaningful differences}, accepting small numerical tolerances, and explicitly forbids re-solving the problem or arguing for an alternative answer. This design follows the LLM-as-judge paradigm established in~\cite{zheng2023mtbench,kim2023prometheus} while tightening it for a domain in which the correct answer is authoritative rather than preferential.

The two benchmarks produce different \emph{views} of the same competence. MCQ mode tests recognition-plus-elimination --- a model that cannot derive the correct answer but can identify implausible options may still succeed. Judge mode tests derivation from scratch: the model must produce a correct answer without the four-option scaffold, and a capable judge then checks it. We report both and contrast them in \cref{sec:results:mvj}.

\subsection{Item Selection for the Benchmarks}
\label{sec:methodology:sampling}

The generation pipeline produces items across five hard difficulty archetypes --- quantitative calculation, assumption sensitivity, conceptual inversion, edge case / boundary and multi-step logic --- sampled with a weight schedule that favors quantitative calculation. Pre-tests showed that items drawn from easier archetypes (terminology precision, consistency checks) are solved nearly uniformly by frontier models and therefore have little discriminative value; we therefore exclude those archetypes from the evaluation pool. Pre-tests further showed that the quantitative-calculation archetype is consistently the most discriminating among the five hard archetypes.

For this study we construct two benchmarks of one hundred items each, disjoint by construction.

\textbf{MCQ benchmark (100 items).} We first generated an internal pool of 200 items spanning the five hard archetypes. From that pool we retain the 100 items on which the \emph{collective accuracy} across the evaluated models was lowest --- the empirically-hardest half. This empirically-hardest-$N$ construction is a standard move in benchmark design~\cite{phan2025hle,naveed2024overview} because static frontier benchmarks saturate quickly under LLM progress: retaining only the items the community's current models most frequently miss preserves discriminative power at the top. The resulting set contains 46~quantitative-calculation items, 17~conceptual-inversion items, 13~assumption-sensitivity items, 13~edge-case items and 11~multi-step-logic items. Cross-model mean accuracy on the retained items ranges from 7.8\,\% (hardest) to 94.1\,\% (easiest).

\textbf{Judge benchmark (100 items).} The Judge benchmark comprises 100 items drawn from a separate pool of approximately 400 items generated by the same pipeline. We restrict the Judge benchmark to the quantitative-calculation archetype because the pre-tests cited above identified it as the most discriminating archetype among the five hard categories, and because the open-ended format most directly rewards genuine derivation on computational tasks. The Judge items are disjoint from the MCQ items.

\subsection{Model Selection}
\label{sec:methodology:models}

We evaluate 50~models spanning eight providers: Anthropic, Google, OpenAI, xAI, DeepSeek, Mistral, Cohere, and an open-weights family of hosted endpoints (Groq, Cerebras) plus a locally-hosted Ollama runtime. Where a provider exposes both a standard and a reasoning-mode variant of the same underlying model (e.g.\ \texttt{claude-opus-4-6} vs.\ \texttt{claude-opus-4-6:thinking}, \texttt{gemini-2.5-pro} vs.\ \texttt{gemini-2.5-pro:thinking}), we include both to support the paired comparison in \cref{sec:discussion:thinking}. Five further models were benchmarked but are excluded from all main tables and figures because adapter-level errors caused every answer to be recorded as unparseable; they are listed in \cref{app:excluded-models}. Table~\ref{tab:models} summarizes the included set. Anthropic Claude is among the evaluated models and is also the tool used to draft this manuscript (see Acknowledgments); we disclose this dual role rather than excluding Claude from the evaluation, because its relative position on the benchmark is of independent scientific interest.

\begin{table}[t]
\centering
\tiny
\begin{tabular}{llrrl}
\toprule
Provider & Model & \$/1M in & \$/1M out & Type \\
\midrule
anthropic & \texttt{claude-opus-4-6:thinking} & 5.000 & 25.000 & thinking \\
openai & \texttt{gpt-5-mini} & 0.125 & 1.000 & dense \\
openai & \texttt{o3:reasoning} & 2.000 & 8.000 & thinking \\
openai & \texttt{o4-mini:reasoning} & 1.100 & 4.400 & thinking \\
cerebras & \texttt{gpt-oss-120b} & --- & --- & open-source \\
google & \texttt{gemini-3.1-pro-preview} & 2.000 & 12.000 & dense \\
google & \texttt{gemini-2.5-pro:thinking} & 1.250 & 10.000 & thinking \\
groq & \texttt{openai/gpt-oss-120b} & 0.150 & 0.600 & open-source \\
openai & \texttt{gpt-5} & 1.250 & 10.000 & dense \\
openai & \texttt{o3-mini:reasoning} & 1.100 & 4.400 & thinking \\
anthropic & \texttt{claude-sonnet-4-6:thinking} & 3.000 & 15.000 & thinking \\
groq & \texttt{openai/gpt-oss-20b} & 0.075 & 0.300 & open-source \\
openai & \texttt{gpt-5-nano} & 0.050 & 0.400 & dense \\
xai & \texttt{grok-3-mini:reasoning} & 0.300 & 0.500 & thinking \\
anthropic & \texttt{claude-sonnet-4-6} & 3.000 & 15.000 & dense \\
google & \texttt{gemini-2.5-pro} & 1.250 & 10.000 & dense \\
google & \texttt{gemma-4-31b-it} & 0.130 & 0.380 & open-source \\
xai & \texttt{grok-3-mini} & 0.300 & 0.500 & dense \\
anthropic & \texttt{claude-opus-4-20250514} & 15.000 & 75.000 & dense \\
cohere & \texttt{command-a-reasoning-08-2025} & 2.500 & 10.000 & thinking \\
google & \texttt{gemini-3-flash-preview} & 0.500 & 3.000 & dense \\
xai & \texttt{grok-4} & 3.000 & 15.000 & dense \\
anthropic & \texttt{claude-opus-4-6} & 5.000 & 25.000 & dense \\
ollama & \texttt{gemma4:latest} & 0.000 & 0.000 & open-source \\
anthropic & \texttt{claude-sonnet-4-20250514:thinking} & 3.000 & 15.000 & thinking \\
google & \texttt{gemma-4-26b-a4b-it} & 0.080 & 0.350 & open-source \\
anthropic & \texttt{claude-sonnet-4-20250514} & 3.000 & 15.000 & dense \\
google & \texttt{gemini-2.5-flash} & 0.150 & 0.600 & dense \\
google & \texttt{gemini-2.5-flash:thinking} & 0.150 & 0.600 & thinking \\
deepseek & \texttt{deepseek-reasoner:reasoning} & 0.280 & 0.420 & thinking \\
google & \texttt{gemini-3.1-flash-lite-preview} & 0.250 & 1.500 & dense \\
groq & \texttt{meta-llama/llama-4-scout-17b-16e-instruct} & 0.110 & 0.340 & open-source \\
openai & \texttt{gpt-5.4} & 2.500 & 15.000 & dense \\
openai & \texttt{gpt-5.2} & 1.750 & 14.000 & dense \\
anthropic & \texttt{claude-haiku-4-5-20251001} & 1.000 & 5.000 & dense \\
openai & \texttt{gpt-4.1} & 2.000 & 8.000 & dense \\
xai & \texttt{grok-3} & 3.000 & 15.000 & dense \\
cohere & \texttt{command-a-03-2025} & 2.500 & 10.000 & dense \\
deepseek & \texttt{deepseek-chat} & 0.280 & 0.420 & dense \\
groq & \texttt{llama-3.3-70b-versatile} & 0.590 & 0.790 & open-source \\
openai & \texttt{gpt-5.4-mini} & 0.750 & 4.500 & dense \\
mistral & \texttt{mistral-large-2512} & 0.500 & 1.500 & dense \\
mistral & \texttt{mistral-large-latest} & 0.500 & 1.500 & dense \\
openai & \texttt{gpt-4.1-mini} & 0.400 & 1.600 & dense \\
openai & \texttt{gpt-4.1-nano} & 0.100 & 0.400 & dense \\
mistral & \texttt{mistral-small-latest} & 0.150 & 0.600 & open-source \\
openai & \texttt{gpt-5.4-nano} & 0.200 & 1.250 & dense \\
groq & \texttt{qwen/qwen3-32b} & 0.290 & 0.590 & open-source \\
groq & \texttt{llama-3.1-8b-instant} & 0.050 & 0.080 & open-source \\
cerebras & \texttt{llama3.1-8b} & 0.100 & 0.100 & open-source \\
\bottomrule
\end{tabular}
\caption{Evaluated models. Per-million-token prices (USD) are the provider's published rates at the time the benchmark was run. Type: \emph{dense} for standard decoder models, \emph{thinking} for reasoning-mode variants, \emph{open-source} for models whose weights are publicly released. Five additional models are excluded due to adapter errors (\cref{app:excluded-models}).}
\label{tab:models}
\end{table}

\subsection{Reproducibility}
\label{sec:methodology:repro}

Two artefacts support replication. First, the 100 MCQ items and 100 Judge items used here, together with every benchmark answer (presented options, shuffle map, raw response, extracted answer, token counts, cost), are published as a browsable interface at \url{https://actubench.de/en/}. Readers can inspect any item and the responses each model gave to it without installing any software. Second, every number in this paper is derived from a single frozen snapshot of the generation database, so that re-running the reporting pipeline reproduces every plot and table deterministically. The generation and benchmark-runner code are available from the author on request.

\section{Results}
\label{sec:results}

All numbers in this section come from the frozen data snapshot described in \cref{sec:methodology:repro} and the 50~included models of \cref{sec:methodology:models}. Throughout this section, ``MCQ accuracy'' refers to accuracy on the 100 MCQ items, and ``Judge accuracy'' to accuracy on the 100 Judge items.

\subsection{Cost--Performance Landscape}
\label{sec:results:pareto}

Figure~\ref{fig:pareto} shows each model as a numbered, provider-coloured point on a cost--accuracy plane. The horizontal axis is the total USD cost of running the full MCQ benchmark (log scale, spanning six orders of magnitude from near-zero for locally-hosted inference up to a few US dollars for the strongest reasoning-mode variants). The vertical axis is MCQ accuracy. Hollow markers are reasoning-mode variants; filled markers are standard models. Model names are listed in the legend under the figure to keep the plot area uncluttered; the Pareto relationship can be read off the scatter by eye.

\begin{figure}[!htbp]
\centering
\input{figures/fig_pareto.pgf}

\vspace{0.3em}
\tiny
\begin{tabular}{@{}rll@{\hspace{1.2em}}rll@{}}
\toprule
\# & Provider & Model & \# & Provider & Model \\
\midrule
1 & \textcolor[HTML]{D97757}{$\circ$} Anth & \texttt{claude-opus-4-6:thinking} & 8 & \textcolor[HTML]{DC50AA}{$\bullet$} Open & \texttt{openai/gpt-oss-120b} \\
11 & \textcolor[HTML]{D97757}{$\circ$} Anth & \texttt{claude-sonnet-4-6:thinking} & 12 & \textcolor[HTML]{DC50AA}{$\bullet$} Open & \texttt{openai/gpt-oss-20b} \\
15 & \textcolor[HTML]{D97757}{$\bullet$} Anth & \texttt{claude-sonnet-4-6} & 24 & \textcolor[HTML]{DC50AA}{$\bullet$} Open & \texttt{gemma4:latest} \\
19 & \textcolor[HTML]{D97757}{$\bullet$} Anth & \texttt{claude-opus-4-20250514} & 32 & \textcolor[HTML]{DC50AA}{$\bullet$} Open & \texttt{meta-llama/llama-4-scout-17b-16e-instruct} \\
23 & \textcolor[HTML]{D97757}{$\bullet$} Anth & \texttt{claude-opus-4-6} & 40 & \textcolor[HTML]{DC50AA}{$\bullet$} Open & \texttt{llama-3.3-70b-versatile} \\
25 & \textcolor[HTML]{D97757}{$\circ$} Anth & \texttt{claude-sonnet-4-20250514:thinking} & 48 & \textcolor[HTML]{DC50AA}{$\bullet$} Open & \texttt{qwen/qwen3-32b} \\
27 & \textcolor[HTML]{D97757}{$\bullet$} Anth & \texttt{claude-sonnet-4-20250514} & 49 & \textcolor[HTML]{DC50AA}{$\bullet$} Open & \texttt{llama-3.1-8b-instant} \\
35 & \textcolor[HTML]{D97757}{$\bullet$} Anth & \texttt{claude-haiku-4-5-20251001} & 50 & \textcolor[HTML]{DC50AA}{$\bullet$} Open & \texttt{llama3.1-8b} \\
20 & \textcolor[HTML]{DAA520}{$\circ$} Cohere & \texttt{command-a-reasoning-08-2025} & 2 & \textcolor[HTML]{10A37F}{$\bullet$} OAI & \texttt{gpt-5-mini} \\
38 & \textcolor[HTML]{DAA520}{$\bullet$} Cohere & \texttt{command-a-03-2025} & 3 & \textcolor[HTML]{10A37F}{$\circ$} OAI & \texttt{o3:reasoning} \\
30 & \textcolor[HTML]{6E46C8}{$\circ$} DSeek & \texttt{deepseek-reasoner:reasoning} & 4 & \textcolor[HTML]{10A37F}{$\circ$} OAI & \texttt{o4-mini:reasoning} \\
39 & \textcolor[HTML]{6E46C8}{$\bullet$} DSeek & \texttt{deepseek-chat} & 9 & \textcolor[HTML]{10A37F}{$\bullet$} OAI & \texttt{gpt-5} \\
6 & \textcolor[HTML]{4285F4}{$\bullet$} Goog & \texttt{gemini-3.1-pro-preview} & 10 & \textcolor[HTML]{10A37F}{$\circ$} OAI & \texttt{o3-mini:reasoning} \\
7 & \textcolor[HTML]{4285F4}{$\circ$} Goog & \texttt{gemini-2.5-pro:thinking} & 13 & \textcolor[HTML]{10A37F}{$\bullet$} OAI & \texttt{gpt-5-nano} \\
16 & \textcolor[HTML]{4285F4}{$\bullet$} Goog & \texttt{gemini-2.5-pro} & 33 & \textcolor[HTML]{10A37F}{$\bullet$} OAI & \texttt{gpt-5.4} \\
17 & \textcolor[HTML]{4285F4}{$\bullet$} Goog & \texttt{gemma-4-31b-it} & 34 & \textcolor[HTML]{10A37F}{$\bullet$} OAI & \texttt{gpt-5.2} \\
21 & \textcolor[HTML]{4285F4}{$\bullet$} Goog & \texttt{gemini-3-flash-preview} & 36 & \textcolor[HTML]{10A37F}{$\bullet$} OAI & \texttt{gpt-4.1} \\
26 & \textcolor[HTML]{4285F4}{$\bullet$} Goog & \texttt{gemma-4-26b-a4b-it} & 41 & \textcolor[HTML]{10A37F}{$\bullet$} OAI & \texttt{gpt-5.4-mini} \\
28 & \textcolor[HTML]{4285F4}{$\bullet$} Goog & \texttt{gemini-2.5-flash} & 44 & \textcolor[HTML]{10A37F}{$\bullet$} OAI & \texttt{gpt-4.1-mini} \\
29 & \textcolor[HTML]{4285F4}{$\circ$} Goog & \texttt{gemini-2.5-flash:thinking} & 45 & \textcolor[HTML]{10A37F}{$\bullet$} OAI & \texttt{gpt-4.1-nano} \\
31 & \textcolor[HTML]{4285F4}{$\bullet$} Goog & \texttt{gemini-3.1-flash-lite-preview} & 47 & \textcolor[HTML]{10A37F}{$\bullet$} OAI & \texttt{gpt-5.4-nano} \\
42 & \textcolor[HTML]{DC3232}{$\bullet$} Mistr & \texttt{mistral-large-2512} & 14 & \textcolor[HTML]{000000}{$\circ$} xAI & \texttt{grok-3-mini:reasoning} \\
43 & \textcolor[HTML]{DC3232}{$\bullet$} Mistr & \texttt{mistral-large-latest} & 18 & \textcolor[HTML]{000000}{$\bullet$} xAI & \texttt{grok-3-mini} \\
46 & \textcolor[HTML]{DC3232}{$\bullet$} Mistr & \texttt{mistral-small-latest} & 22 & \textcolor[HTML]{000000}{$\bullet$} xAI & \texttt{grok-4} \\
5 & \textcolor[HTML]{DC50AA}{$\bullet$} Open & \texttt{gpt-oss-120b} & 37 & \textcolor[HTML]{000000}{$\bullet$} xAI & \texttt{grok-3} \\
\bottomrule
\end{tabular}

\caption{Cost--performance landscape on the MCQ benchmark. Each point is one evaluated model, coloured by provider; filled markers are standard models, hollow markers are reasoning-mode variants. Numbered points are identified in the legend above.}
\label{fig:pareto}
\end{figure}

Three features of the landscape stand out. First, the accuracy ceiling is \emph{crowded}: four models --- \texttt{claude-opus-4-6:thinking}, \texttt{gpt-5-mini}, \texttt{o3:reasoning} and \texttt{o4-mini:reasoning} --- reach the top MCQ accuracy of 98\,\% (98/100 items correct), yet their total-run costs span a full order of magnitude, from about nine US cents up to over \$1.50 for the same accuracy. On hard but structurally closed multiple-choice items, per-answer frontier performance can be obtained much more cheaply from the right non-flagship model than from the flagship.

Second, two models strictly dominate at the zero-cost end of the scale: a hosted open-weights 120B-parameter model from Cerebras at 97\,\% MCQ accuracy at an extremely low cost (effectively a fraction of a US cent for the full 100-item run), and a locally-hosted Gemma~4 model running on consumer hardware at 85\,\% accuracy and zero marginal cost. The Cerebras endpoint lies within one item of the top of the leaderboard; the Ollama case shows that a medium-sized open-weights model running on consumer hardware achieves accuracy within 13~percentage points of the best paid option.

Third, the space between the Pareto-dominant points at the zero-cost end and the top-accuracy cluster is thin. A reader might be tempted to interpret this as reasoning-mode inference being wasteful, but that overstates the case: the MCQ task as posed (recognition plus elimination) is not the hardest test these models can be given, and the compression at the ceiling is partly an artefact of it. The Judge-mode results below (\cref{sec:results:mvj}) decompress the top of the scale.

\subsection{Accuracy by Actuarial Sector}
\label{sec:results:sectors}

Each item in the MCQ benchmark is tagged by the auxiliary agent (\cref{sec:pipeline:aux}) with zero or more actuarial-sector labels: life, non-life, AFIR-ERM (actuarial approach for financial risks and enterprise risk management), ADS-AI (actuarial data science and AI), health, and pension. Labels are not mutually exclusive. Table~\ref{tab:sectors} reports the mean accuracy per sector, pooled over all 50 included models.

\begin{table}[t]
\centering
\begin{tabular}{lrr}
\toprule
Sector & \#Items & Mean accuracy \\
\midrule
ADS-AI & 23 & 0.870 \\
Health & 2 & 0.790 \\
Life & 24 & 0.777 \\
Pension & 5 & 0.756 \\
AFIR-ERM & 41 & 0.748 \\
Non-life & 12 & 0.733 \\
\bottomrule
\end{tabular}
\caption{Mean accuracy per actuarial sector on the MCQ benchmark, pooled over the 50 included models. Item counts per sector are uneven because each item may carry multiple labels and because the IAA syllabus is itself unevenly weighted across sectors. Small-sample sectors (health, pension) should be read with a correspondingly wider confidence band.}
\label{tab:sectors}
\end{table}

At the top of the ranking the sector-level accuracies are near-uniform: the five strongest models score within a couple of items of 100\,\% on every sector. Differentiation appears in the full distribution over all 50 models. Mean accuracy is highest on ADS-AI and lowest on non-life; the most statistically robust figure is AFIR-ERM, which carries the largest item count inside the benchmark and is the best estimate of how frontier and near-frontier LLMs handle advanced enterprise-risk and capital-management content. The uneven sector counts per MCQ benchmark are a direct artefact of the sampling over IAA learning objectives; a future study that rebalances the sample could sharpen the smaller-sample estimates.

\subsection{MCQ vs.\ Judge Divergence}
\label{sec:results:mvj}

\begin{figure}[!htbp]
\centering
\input{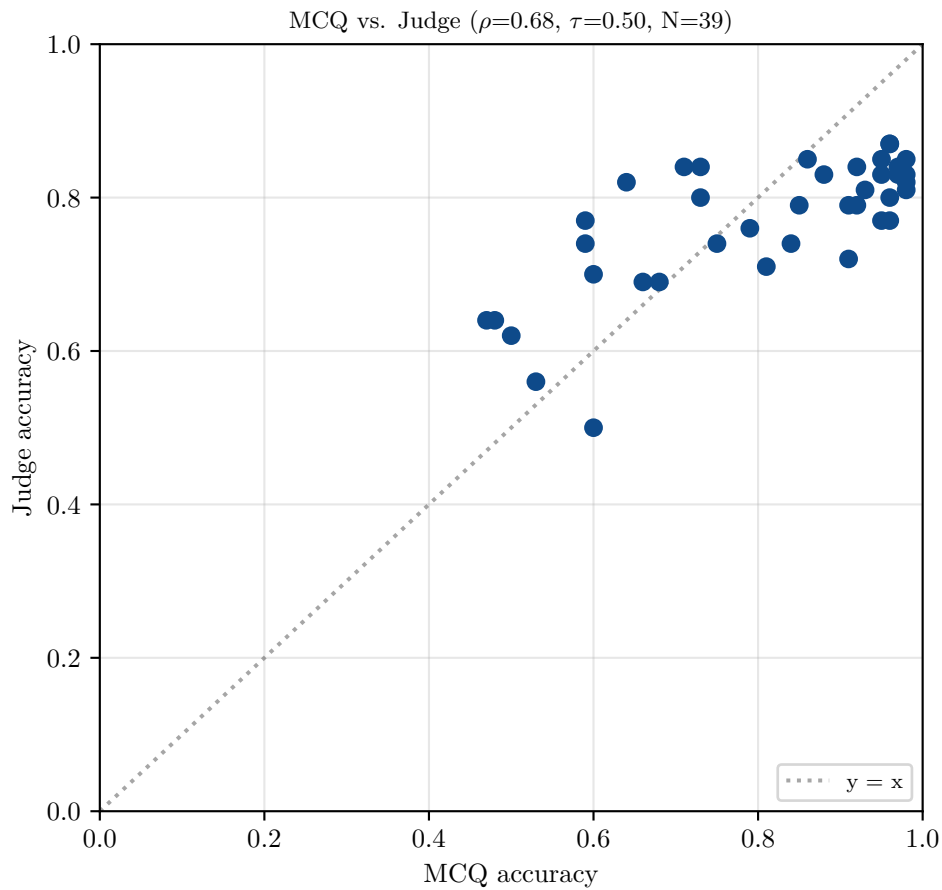}
\caption{Per-model accuracy on the 100-item MCQ benchmark (x-axis) versus the 100-item Judge benchmark (y-axis). The dotted line is $y=x$; points below the diagonal are models that lose accuracy when the four-option scaffold is removed. Spearman $\rho$ and Kendall $\tau$ over the $N=39$ commonly evaluated models are reported in the figure title.}
\label{fig:mcq_vs_judge}
\end{figure}

To contrast the two evaluation modes we plot each model's MCQ accuracy (on the 100-item MCQ benchmark) against its Judge accuracy (on the 100-item Judge benchmark) in \cref{fig:mcq_vs_judge}. The two item sets are disjoint by construction but both come from the same generation pipeline, with the Judge pool restricted to the most discriminating difficulty archetype (\cref{sec:methodology:sampling}).

The picture is more nuanced than an informal ``Judge is harder'' hypothesis would suggest. At the top of the MCQ ranking the two benchmarks diverge sharply: the best MCQ score is 98\,\%, the best Judge score 87\,\%, and several frontier models lose on the order of 10--15~pp when the four-option scaffold is removed. For lower-tier models the sign of the gap frequently reverses: a number of smaller or cheaper models gain 10--20~pp on Judge relative to their MCQ score. When a model cannot reliably solve the underlying computation, MCQ-style elimination exposes that more cleanly than the Judge's looser numerical-tolerance matching does, so the Judge can end up being the more forgiving instrument at the low end of the scale.

The finding restated: the MCQ scaffold inflates the performance ceiling. At the top of the benchmark, where every strong model is near 100\,\% on MCQ, Judge mode produces a more spread-out ranking with a ceiling closer to 87\,\%; at the bottom of the benchmark, the judge's tolerant matching can be more forgiving than MCQ's one-letter-only scoring. The rank correlations reported in \cref{fig:mcq_vs_judge} reflect exactly this: moderate agreement, not identity, between the two benchmarks' orderings. Practically, this means MCQ-only actuarial benchmarks should not be read as a ranking of genuine reasoning capacity on open-ended problems. Judge-mode evaluation is a sharper instrument near the frontier and a differently-shaped instrument further down; both views are needed.

\section{Discussion}
\label{sec:discussion}

\subsection{Reasoning-Mode vs.\ Standard Variants}
\label{sec:discussion:thinking}

\begin{figure}[!htbp]
\centering
\input{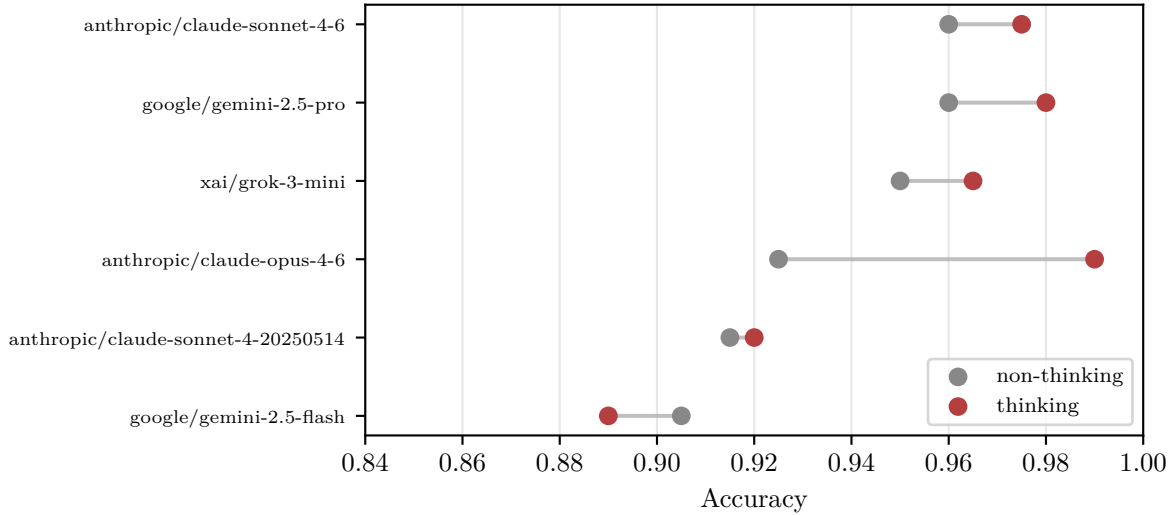}
\caption{Accuracy of six reasoning-mode (red) variants against their standard-decoder siblings (grey), on the MCQ benchmark. The dumbbell is ordered by non-thinking accuracy. The x-axis is zoomed to the region where the paired variants actually sit. Differences are typically small (one to four percentage points); the Anthropic Opus pair is the outlier at +13~pp.}
\label{fig:thinking}
\end{figure}

Six provider/model pairs in our evaluation expose both a standard decoder variant and an explicit reasoning-mode (``thinking'') variant built on the same base model. Figure~\ref{fig:thinking} shows the paired comparison on the MCQ benchmark. The aggregate picture is modest: across the six pairs, the mean accuracy gain from enabling the reasoning mode is about three and a half percentage points, at an average inference-cost multiple around two and a half. The distribution is skewed: one pair --- \texttt{claude-opus-4-6} versus \texttt{claude-opus-4-6:thinking} --- moves from 85\,\% to 98\,\% accuracy (+13~pp) at roughly 4.7$\times$ the cost; most other pairs move by one to four percentage points. One pair shows a slight regression (\texttt{gemini-2.5-flash:thinking} scores 2~pp below its non-thinking sibling on this set).

Two implications follow. First, reasoning mode is not free performance: on a well-constructed hard-MCQ benchmark it delivers a few percentage points of accuracy per factor of two-to-three in cost, with high variance across model families. Second, the cost-aware reader of \cref{sec:results:pareto} should not equate ``reasoning mode'' with ``Pareto-dominant''; the two Pareto-dominant models at the zero-cost end of the scale in our study are both \emph{standard} open-weights models (\texttt{gpt-oss-120b} on Cerebras and Gemma~4 via Ollama). Where reasoning mode does dominate is at the frontier ceiling: five of the top six scores on the MCQ benchmark come from reasoning-mode variants.

\subsection{Agent-C Repair Statistics}
\label{sec:discussion:agentc}

Of the items Agent~C reviewed during generation of the MCQ benchmark pool, roughly three in five were flagged on first pass at the item level (item stem + correct answer) and just under half were flagged at the distractor-ensemble level. These rates are substantially higher than any self-verification rate we observed when the same adapter that drafted an item also verified it in preliminary experiments --- consistent with the literature on the verification-generation asymmetry~\cite{du2023multiagent_debate,zheng2023mtbench}. An in-distribution generator is a poor auditor of its own output.

Repair is largely effective. Of the items flagged at the item level, roughly three-quarters pass the re-check after a single-shot Agent~A repair and proceed normally; the remainder --- around one in six items overall --- are persisted with a failure flag. The distractor-level loop behaves comparably: most flagged distractor sets are repaired on a single Agent~B pass, a minority persist the failure flag. In aggregate, a pipeline that \emph{discarded} repair failures rather than marking and keeping them would lose a non-trivial fraction of its output. The design decision to keep-and-flag rather than discard was made for dataset completeness and is re-visitable for studies that require perfectly verified subsets only.

\subsection{Implications for Practical Actuarial LLM Use}
\label{sec:discussion:implications}

Three practical conclusions follow from our results. For \emph{multiple-choice or structured decision support}, a small or mid-tier general-purpose model (for example \texttt{gpt-5-mini} at roughly ten US cents per hundred answers, or a hosted open-weights endpoint at near-zero cost) delivers accuracy within two percentage points of the flagship reasoning variants on the hardest actuarial items we could construct. The extra money spent on the most expensive reasoning-mode models buys little additional accuracy on this task format.

For \emph{open-ended reasoning and derivation}, the picture reverses: the MCQ ranking compresses models at the top, while Judge-mode scoring reveals a roughly 15\,pp gap between the strongest frontier reasoning models (about 85\,\% Judge accuracy) and merely strong non-reasoning models (about 70\,\%). When the task actually requires deriving an answer rather than selecting one, reasoning-mode or frontier-generation models are cost-justified.

For \emph{local deployment}, the observation that Gemma~4 running via Ollama on a single consumer GPU reaches 85\,\% accuracy on the MCQ benchmark is material. A team that prefers --- for data-governance, cost-control or latency reasons --- to run inference locally is not paying a large accuracy penalty on structured actuarial knowledge; on open-ended reasoning the penalty is larger, and a hybrid architecture (local for MCQ-style classification, hosted frontier for derivation) is likely the right trade-off.

\section{Limitations and Future Work}
\label{sec:limitations}

Five limitations of the present study deserve explicit note.

\textbf{Statistical power.} Each of our two benchmarks contains 100 items. A 5-point accuracy difference between two models at 90\,\% mean accuracy has a Wilson 95\,\% confidence interval of roughly $\pm$6~pp, which means small rank differences in the middle of our tables are not statistically decisive. The MCQ-benchmark design choice of empirically-hardest-100 trades sample size for discriminative power at the ceiling; a larger study that re-ran the pipeline to produce, say, 400 items per benchmark would tighten the mid-table estimates at proportionately higher evaluation cost.

\textbf{Factual grounding via Wikipedia.} The generation pipeline (\cref{sec:pipeline:agenta}) uses Wikipedia as its sole factual anchor. This is a deliberate choice: Wikipedia is reproducibly fetchable, well-understood as a distribution, and its extracts can be versioned. But it under-covers two important regions of the IAA syllabus: non-English regulatory specifics (Solvency II technical standards, national actuarial standards of practice) and highly specialized practice literature (professional-body notes, consulting-firm methodology papers). Items that would require those sources either degrade to Wikipedia-adjacent paraphrase or are not generated at all.

\textbf{Language.} The current generation pipeline and both benchmarks are English-only; Wikipedia is accessed on the English edition. A German-language ActuBench would need a German note-summarization adapter and would produce a different distribution of items on the same syllabus. We consider this the most concrete near-term extension, given that the DAV examination ecosystem the work motivates is itself German-language.

\textbf{Judge-model bias.} The LLM-as-judge design imports known judge-side biases: position, verbosity, and stylistic preferences of the judge model can influence verdicts independently of correctness~\cite{wang2023unfair}. For closed-ended actuarial items with ground-truth answers the exposure is smaller than for preference-style evaluation, but it is not zero. A robustness check that varies the judge model and reports inter-judge agreement would strengthen the Judge-mode claims of~\cref{sec:results:mvj}; we regard this as the most important methodological follow-up.

\textbf{Contamination risk.} Although our pipeline can generate fresh items on demand and the item subset we release is curated, the generation pipeline itself is LLM-driven and its prompts and Wikipedia sources are public. Any generated item released publicly risks eventual inclusion in future pre-training corpora. The live-website release model of~\cref{sec:methodology:repro} slows but does not prevent this. A long-running ActuBench evaluation would rotate its public subset periodically.

\section{Conclusion}
\label{sec:conclusion}

We have presented ActuBench, a multi-agent LLM pipeline for the automated generation and evaluation of advanced actuarial assessment items. The pipeline's principal methodological move is an \emph{independent verifier agent}: separate adapters draft, construct distractors, and verify; each verdict is structured and auditable; one-shot repair loops make a substantial fraction of flagged items usable without human intervention. The verifier catches first-pass problems in a majority of drafted items at the stem level and in close to half at the distractor level, most of which are subsequently repaired by the one-shot loop.

We evaluated fifty language models spanning eight providers on two complementary benchmarks --- 100 empirically hardest multiple-choice items and 100 open-ended items scored by an LLM judge. Three findings are our headline empirical contribution. At the top of the MCQ benchmark, four models share the ceiling at 98\,\% accuracy while their inference costs span an order of magnitude: flagship accuracy on closed-form items is cheaper than it looks. A hosted open-weights 120B model and a locally-run Gemma 4 hold two Pareto-dominant positions at near-zero marginal cost. And MCQ-mode and Judge-mode do not rank models identically: removing the four-option scaffold shifts the benchmark's ceiling downward by about 10--15~pp and reveals cleaner differentiation among the strongest reasoning-mode models.

The items and per-model responses are browsable at \url{https://actubench.de/en/}; the pipeline code is available from the author on request. We hope the combination of independent verification on the generation side, a ground-truth LLM-as-judge mode on the evaluation side, and a curated public item subset makes ActuBench useful as both a practical evaluation tool for the actuarial community and an empirical reference for the broader LLM-benchmark literature.

\printbibliography

\appendix
\section{Key Prompts}
\label{app:prompts}

This appendix reproduces the core prompts used by Agents~A, B and~C, and by the auxiliary agent. All prompts request JSON-only output; validation and caching are handled by the pipeline. For space, we quote the prompt instructions in condensed form; the full templates, including structured output schemas, are available from the author on request.

\textbf{Agent~A --- item drafting.} Given the IAA learning objective, the sampled difficulty archetype, and the Wikipedia notes assembled by the auxiliary agent, Agent~A is instructed to produce the question stem, the correct answer, and its rationale. Hard constraints: English only; a single best answer under all reasonable interpretations; explicit statement of any assumption needed for uniqueness; no reliance on parametric knowledge that cannot be traced back to the notes. The difficulty archetype is passed as a short natural-language description (e.g., ``Quantitative calculation: multi-step computations where each sub-step admits a characteristic error'') so the model calibrates style and depth accordingly.

\textbf{Agent~B --- distractor construction.} Agent~B is shown the stem, the correct answer with its rationale, and the Wikipedia notes. It is asked for three distractors, each with a short free-text rationale stating the specific misconception or computational slip the distractor is meant to probe, and each satisfying individual plausibility, collective distinctness, and individual incorrectness under the stem's stated assumptions.

\textbf{Agent~C --- item verification.} Agent~C receives the learning objective, the Wikipedia notes, the stem, the correct answer, and the rationale, but \emph{not} the identity of Agent~A or the drafting prompt. It returns a structured verdict \{ok: bool, issues: [\{type: factual\_error\,$|$\,ambiguity\,$|$\,underspecification\,$|$\,difficulty\_misalignment, description: str\}]\}. On ok=false, Agent~A's repair stage is re-invoked with the full issue report.

\textbf{Agent~C --- distractor verification.} Agent~C receives all four options with their rationales. It returns the same verdict schema, with issue types specialized to distractor design: duplicate\_with\_A, duplicate\_with\_other, not\_plausible, inadvertently\_correct.

\textbf{Auxiliary agent --- Wikipedia notes.} Given a Wikipedia extract, the auxiliary agent produces a fixed bullet-list schema grouping the content into named facts and their provenance. The same adapter is re-used for topic-label assignment at the end of the pipeline.

\textbf{Judge-mode prompts.} The evaluatee sees a minimal system prompt (``You are taking an actuarial exam. Answer as accurately and concisely as possible. Show your reasoning briefly, then state your final answer clearly.'') and the bare stem. The judge sees the original question, the evaluatee's free-text response, and the known correct answer, and returns a JSON verdict comprising the extracted final answer, a short reasoning note, a yes/no correctness decision, and a 0--100 confidence score.

\section{Example Items}
\label{app:examples}

Two example items from the MCQ benchmark are shown below. Option~A is always the correct answer in the stored item; at evaluation time the four options are shuffled per run and the shuffle map is persisted (\cref{sec:methodology:mcq}).

\subsection*{Example 1 --- Quantitative Calculation}

\emph{Learning objective 4.2.2:} Apply techniques for creating new distributions: multiplication by a constant, raising to a power.

\textbf{Stem.} A random variable $X$ follows a log-normal distribution with parameters $\mu = 1.2$ and $\sigma^2 = 0.09$, meaning $\ln(X) \sim \mathrm{Normal}(\mu, \sigma^2)$. Define a new random variable $Y = 4X^2$. Given $\ln(4) \approx 1.386$, what are the parameters $(\mu_Y, \sigma_Y^2)$ of the log-normal distribution of~$Y$?

\begin{itemize}\setlength{\itemsep}{0pt}\setlength{\parskip}{0pt}
  \item \textbf{A} (correct): $\mu_Y = 3.786,\ \sigma_Y^2 = 0.36$.
  \item B: $\mu_Y = 2.4,\ \sigma_Y^2 = 0.36$.
  \item C: $\mu_Y = 3.786,\ \sigma_Y^2 = 0.18$.
  \item D: $\mu_Y = 5.172,\ \sigma_Y^2 = 0.36$.
\end{itemize}

\textbf{Rationale.} Since $\ln(X) \sim \mathrm{Normal}(1.2, 0.09)$, we have $\ln(Y) = \ln(4) + 2\ln(X) \sim \mathrm{Normal}(\ln(4) + 2\cdot 1.2,\ 4 \cdot 0.09) = \mathrm{Normal}(3.786, 0.36)$. Distractors B and~D test incorrect handling of the additive constant; C tests incorrect variance scaling (multiplying $\sigma^2$ by 2 rather than by~4).

\subsection*{Example 2 --- Conceptual Inversion}

\emph{Learning objective 2.1.5:} Apply the term structure of interest rates to modelling cash flows.

\textbf{Stem.} An analyst values a single, certain cash flow using standard present-value discounting with the appropriate spot rate from the government yield curve (no embedded options, no credit adjustments). She finds that the present value of a \$50{,}000 payment due in 18~months is \$50{,}750. What condition must exist in the current interest-rate environment for this result to be internally consistent with standard discounting theory?

\begin{itemize}\setlength{\itemsep}{0pt}\setlength{\parskip}{0pt}
  \item \textbf{A} (correct): The 18-month spot rate must be negative.
  \item B: The 18-month spot rate must be positive, because a positive rate naturally produces a slight premium over the face amount.
  \item C: The yield curve must be inverted, meaning long-term spot rates are lower than short-term spot rates.
  \item D: The 18-month implied forward rate must be negative, but the 18-month spot rate may remain positive.
\end{itemize}

\textbf{Rationale.} Under standard discounting, $\mathrm{PV} = \mathrm{FV}/(1+r)^t$. Here $\mathrm{PV} > \mathrm{FV}$, so $(1+r)^{1.5} < 1$, which requires $r < 0$. Distractor~B inverts the direction of the PV-FV relationship; C confuses the shape of the curve with the sign of the level; D confuses forward and spot rates.

\section{Excluded Models}
\label{app:excluded-models}

Five models were run in the benchmark but are excluded from all main tables and figures because adapter-level errors caused every answer to be recorded as unparseable (zero accuracy on all items of the MCQ pool):

\begin{itemize}\setlength{\itemsep}{0pt}\setlength{\parskip}{0pt}
  \item \texttt{anthropic/claude-opus-4-20250514:thinking}
  \item \texttt{groq/qwen/qwq-32b:reasoning}
  \item \texttt{groq/deepseek-r1-distill-llama-70b:reasoning}
  \item \texttt{mistral/magistral-medium-2509:reasoning}
  \item \texttt{mistral/ministral-3-8b-2512}
\end{itemize}

In all five cases the zero accuracy reflects the adapter's inability to retrieve a parseable letter from the response (thinking-mode outputs exceeding the token budget, or unexpected response envelopes). These are adapter-implementation issues, not capability measurements; a future pipeline update should either fix the adapters or skip the affected models cleanly.

\section*{Acknowledgments --- LLM-assistance disclosure}

The manuscript was drafted with the assistance of Anthropic Claude (Opus 4.7, 1M-context). All research design, empirical interpretations and final text decisions are the responsibility of the author. Claude is additionally one of the evaluated models in this study (\cref{sec:methodology:models}); we disclose this dual role rather than excluding Claude from the evaluation, because its relative position on the benchmark is of independent scientific interest.

\end{document}